\documentclass[letterpaper, 10 pt, conference]{ieeeconf}  %

\IEEEoverridecommandlockouts
\usepackage{graphicx} 
\usepackage{xcolor}
\usepackage{biblatex}
\usepackage{subcaption}
\usepackage{scalerel}

\usepackage{tabularx}
\title{\LARGE \bf
A short methodological review on social robot navigation benchmarking
}

\author{Pranup Chhetri$^{1}$, Alejandro Torrej\'on$^{2}$, Sergio Eslava$^{2}$, Luis J. Manso$^{1}$
\thanks{$^{1}$Pranup Chhetri and Luis J. Manso are with the Department of Artificial Intelligence and Robotics, Aston University, Aston Triangle, B47ET Birmingham, United Kingdom ({\tt\small l.manso@aston.ac.uk})}%
\thanks{$^{2}$Alejandro Torrej\'on and Sergio Eslava are with the Department of Computer and Communication Technology,
Universidad de Extremadura, Avd. de la Universidad, 10001
C\'aceres, Extremadura ({\tt\small atorrejon@unex.es, sergioeslava@unex.es})}%
}

\let\labelindent\relax
\usepackage{enumitem}

\begin{document}

\maketitle
\thispagestyle{empty}
\pagestyle{empty}

\begin{abstract}
Social Robot Navigation is the skill that allows robots to move efficiently in human-populated environments while ensuring safety, comfort, and trust. Unlike other areas of research, the scientific community has not yet achieved an agreement on how Social Robot Navigation should be benchmarked. This is notably important, as the lack of a \textit{de facto} standard to benchmark Social Robot Navigation can hinder the progress of the field and may lead to contradicting conclusions. Motivated by this gap, we contribute with a short review focused exclusively on benchmarking trends in the period from January 2020 to July 2025. Of the 130 papers identified by our search using IEEE Xplore, we analysed the 85 papers that met the criteria of the review. This review addresses the metrics used in the literature for benchmarking purposes, the algorithms employed in such benchmarks, the use of human surveys for benchmarking, and how conclusions are drawn from the benchmarking results, when applicable.
\end{abstract}

\section{INTRODUCTION}
Benchmarking is key to providing evidence of the effectiveness of algorithms under equitable conditions.
For instance, in object detection, metrics such as IoU~\cite{lin2014microsoft} are essentially taken as a standard, allowing objective comparisons between different algorithms.
Unlike computer vision, the Social Robot Navigation (SocNav) community has not yet reached an agreement on how SocNav algorithms should be benchmarked.
This is, arguably, not due to lack of interest but due to the subjective nature of what \textit{quality} means in the context of SocNav, and to the inherent complexity of human-robot interactions.
Although, theoretically, in-situ surveys constitute one of the benchmarking approaches with strongest support from the research community, in practice, the time and operational costs of running significant and reproducible surveys frequently make them unfeasibly expensive~\cite{francis2025principles}.

\par
The absence of a \textit{de facto} standard benchmark for SocNav can hinder the progress of the field, as research can reach contradictory or flawed conclusions.
Moreover, this absence can generate confusion for SocNav researchers, as it can be challenging to establish which algorithms their contributions should be compared against, which metrics should be used, and how to make a final decision whenever the metrics point in different directions.
In light of the aforementioned situation, this review addresses the following research questions related to SocNav benchmarking:
\begin{enumerate}
  \item What metrics and algorithms are used?
  \item How frequent are surveys involving human raters?
  \item How are benchmarking results interpreted?
\end{enumerate}


\begin{figure}[t]
    \centering
    \includegraphics[width=0.85\linewidth]{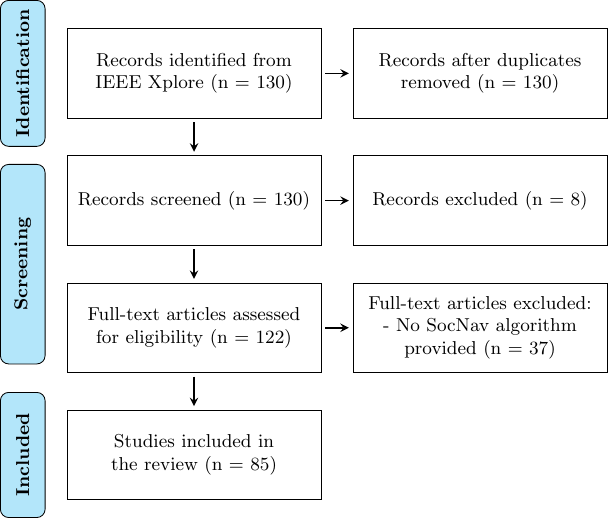}
    \caption{PRISMA flow diagram of the process followed.}
    \label{fig:prisma}
\end{figure}

\section{METHODOLOGY} \label{method}
Given that the purpose of this paper is to provide an overview of the latest trends in benchmarking SocNav algorithms rather than a fully-comprehensive review, the literature search was conducted exclusively on IEEE Xplore.
Nevertheless, the methodology followed a structured approach to ensure reproducibility.

The search query used was \textit{("social robot navigation" OR "social navigation")}, and it was applied to the full text and metadata of the articles to maximise recall.
Also in line with our goal of focusing on the current trends, the search was constrained to works published between January 2020 and July 2025, with all document types included (\textit{i.e.}, conference and journal papers).
The 130 papers returned by the search query were subjected to a two-stage screening process, as depicted in Fig.~\ref{fig:prisma}:
\begin{itemize}[leftmargin=*]
    \item Topic relevance: Despite matching the search query, some of the papers retrieved addressed SocNav tangentially and did not have it as their central focus. This constituted an exclusion criterion. Examples include works referring to SocNav as a necessary skill or an application area (\textit{e.g.}, trajectory prediction). As a result, eight papers were discarded after manual inspection, leaving  122 papers.
    \item Algorithmic contribution: Of the remaining papers, 44 were excluded because they did not propose a \textbf{navigation algorithm}. After this second exclusion criterion, 85 papers were retained for review ($65\%$ of the total retrieved).
\end{itemize}
The screening was evenly distributed among the authors of the review, each paper being screened by two different members of the team to ensure consistent application of the criteria.
In case of doubt or conflict, the team discussed the classification collectively.

\begin{figure}
    \centering
    \includegraphics[width=\linewidth, clip, trim=0 0 0 0]{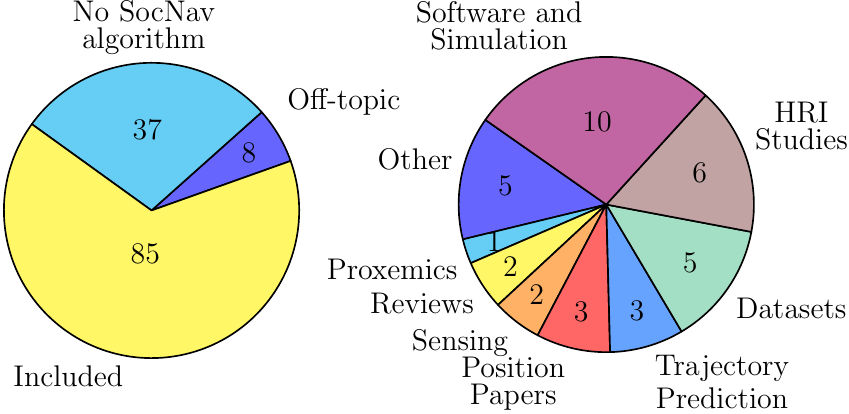}
    \caption{On the left, an overview of the overall screening decisions. On the right, the distribution of the rationale for the application of the second exclusion criterion.}
    \label{fig:pies}
\end{figure}

The papers excluded after applying the second exclusion criterion were: HRI studies~(6), datasets-only papers~(5), position papers~(3), reviews~(2), or focused on other algorithmic topics other than robot control~(44). The papers in this last category were categorised as: software \& simulation~(10), human trajectory prediction~(3), sensing~(3), proxemics~(1), or other algorithmic questions (5, with only 1 item per category).
The distribution of the reasons why papers were excluded is shown in Fig.~\ref{fig:pies}.
\par
Each of the final 85 papers that passed the criteria was reviewed to extract:
\begin{itemize}[leftmargin=*]
    \item the \textbf{quantitative} navigation metrics used,
    \item the algorithms used for \textbf{for quantitative} benchmarking,
    \item the use of surveying \textbf{for benchmarking}, and
    \item how \textbf{conclusions} were extracted from the benchmarking process.
\end{itemize}

\subsection{Further clarifications regarding metric selection}
Metrics accounting for a phenomenon or its absence were considered equivalent (\textit{e.g.}, ratio of missions with and without a collision). The same applied to metrics differing only by a threshold, such as Space Compliance ($SC$) and the Intimate Space Compliance.
\par
Some works used weighted averages of metrics to circumvent the choice of a single metric.
Although this can simplify comparisons, the weights were arbitrary in all instances encountered.
In those cases, we considered the components rather than their weighted average.
\par
Finally, metrics not directly related to navigation, such as perception or communication cues, were excluded.

\subsection{Further clarifications regarding algorithm selection}
Comparisons against pedestrian movement in a dataset or against a manually operated robot were also disregarded, as such comparisons depend on the specific pedestrian or operator's skills and do not establish a measurable property of a trajectory or algorithm.
Similarly, ablations of a proposed algorithm, and algorithms varying only in parameters (\textit{e.g.}, the profile of social force models), were also considered as the same algorithm.

\section{RESULTS}
\subsection*{RQ1) What metrics and algorithms are used?}
Of the $85$ papers reviewed, $53$ provided \textbf{quantitative metrics} for comparison against 3rd-party algorithms ($62.4\%$), $17~(20\%)$ reported quantitative results using metrics but no comparison against other algorithms, and $15~(17.6\%)$ did not provide any quantitative results on navigation performance (some of these relied to human surveys).
Overall, $66$ papers $(77.6\%)$ used non-social metrics, $46~(54.1\%)$ reported social metrics,  and $40~(47.1\%)$ used both types.
\par
As for the rationale for metric selection, out of the $69$ papers using any metric, $36$ papers $(52.2\%)$ did not provide any rationale, $19~(27.5\%)$ justified their choice (\textit{e.g.}, based on good coverage of desired features), and $14~(20.3\%)$ referred to the metrics selected being widely used, seen in previous literature, or used in a paper the authors built upon.
\par
Regarding the rationale behind selecting algorithms for benchmarking, of the $53$ papers comparing their performance against 3rd-party baselines, $4~(7.3\%)$ chose based on similarity, $8~(15.1\%)$ based on the papers they improved on, $7~(13.2\%)$ based on popularity, $9~(16.4\%)$ based on the papers being state-of-the-art, $3~(5.7\%)$ based on subjective perception of quality, $3~(5.7\%)$ followed choices made in previous literature, and $20~(35.8\%)$ did not specify any rationale.
\par
\begin{figure*}[t]
    \centering
    \vstretch{.9525}{\includegraphics[width=\linewidth]{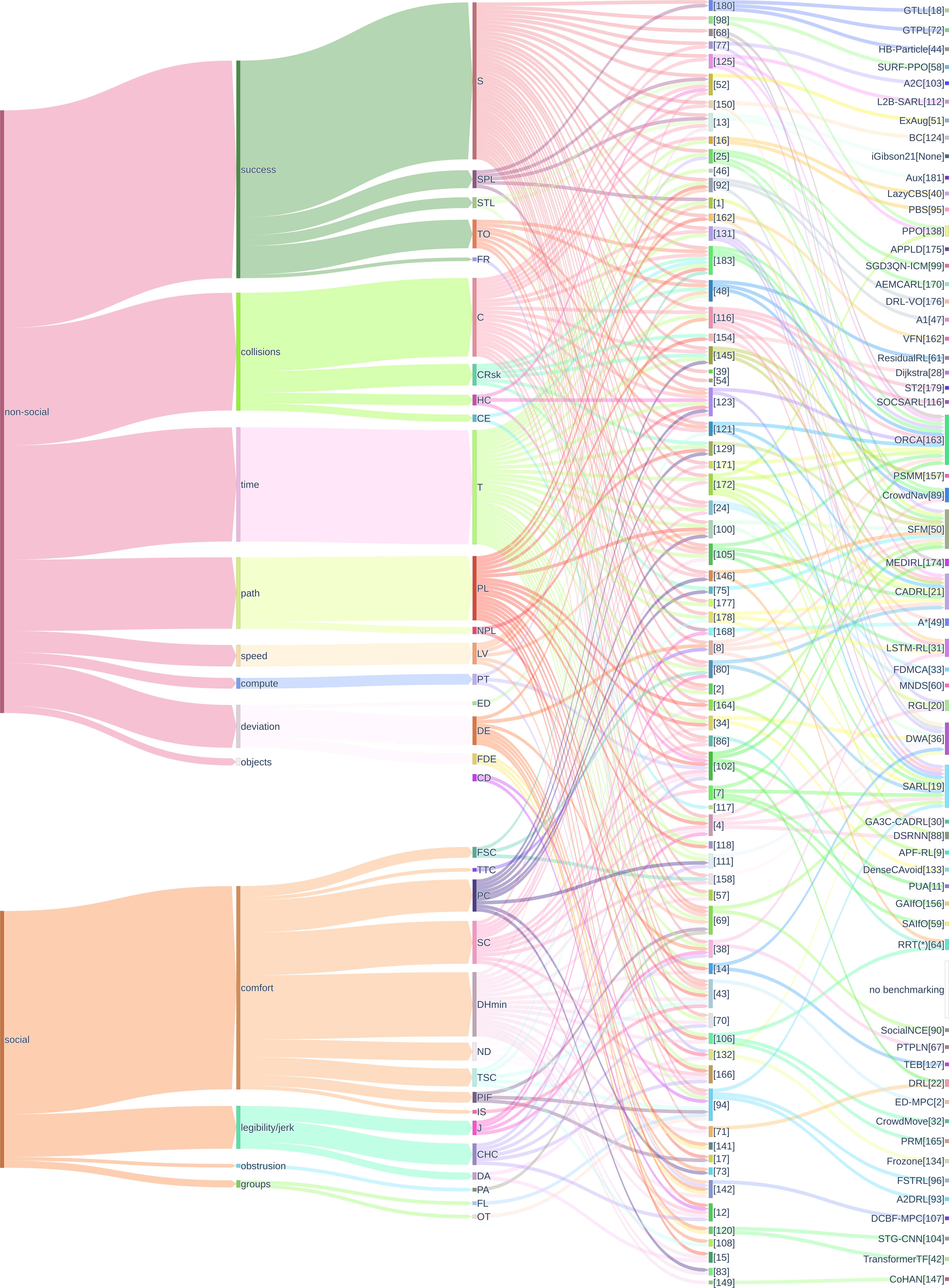}}
    \caption{Sankey diagram showing, left to right, a categorisation of metrics in social \textit{vs.} non-social, the general variables measured by the metrics, the metrics found in the review, the works reviewed in this paper, and the baselines they used.}
    \label{fig:sankey}
\end{figure*}

The metrics used for comparison, after the grouping as described in Sec~\ref{method}, are shown in Table~\ref{tab:metrics}.
The algorithms used for comparison are shown in Fig.~\ref{fig:sankey}, along with the papers and the metrics used in the comparisons.

\setlength{\tabcolsep}{2pt}
\begin{table}[h]
\centering
\resizebox{\columnwidth}{!}{%
\begin{tabularx}{1.05\columnwidth}{|l|X|l|l|}
\hline
\textbf{Short} & \textbf{Definition} & Variable & \textbf{Paper} \\
\hline
$C$ & Collision. &  collision & \cite{silva_towards_2022} \\
$CD$ & Clearing Distance from the robot to the closest object. &objects& \cite{lu_group-aware_2025} \\
$CE$ & Collision Energy. & collision & \cite{zhu_confidence-aware_2025} \\
$CHC$ & Cumulative Heading Changes. &legibility&\cite{ma2024spatiotemporal} \\
$C_{risk}$ & Estimated risk of collision. &collision& \cite{kivrak_multilevel_2020} \\
$DA$ & Deviation angle. &deviation& \cite{sathyamoorthy_comet_2022} \\
$DE$ & Error between the target and actual trajectory.& deviation& \cite{shamsah_real-time_2024} \\
$DH_{min}$ & Minimum Distance to Humans. & comfort & \cite{francis2025principles} \\
$ED$ & End Displacement over a threshold. &deviation& \cite{han_dr-mpc_2025} \\
$FDE$ & Final Displacement Error. &deviation& \cite{shamsah_real-time_2024} \\
$FL$ & Instances where a robot follows a group of pedestrians. &group& \cite{lu_group-aware_2025} \\
$FR$ & Freezing Robot. &success& \cite{sathyamoorthy_comet_2022} \\
$FSC$ & Full Space Compliance over a trajectory. & comfort & \cite{zhu_confidence-aware_2025} \\
$HC$ & Human Collisions. &collision & \cite{pohland_stranger_2024} \\
$IS$ & Robot's speed when breaking Space Compliance. & comfort & \cite{golchoubian_uncertainty-aware_2024} \\
$J$ & The average change in acceleration per unit time. &legibility& \cite{ansari_exploring_2023} \\
$LV$ & Linear Velocity. &speed& \cite{katyal_learning_2022} \\
$ND$ & Number of Discomfort instances. & comfort &\cite{mustafa_context_2024} \\
$NPL$ & Normalised Path Length --- ratio between the path's length and the distance between the start and goal locations. &path& \cite{sathyamoorthy_comet_2022} \\
$NT$ & Navigation Time.&time& \cite{pohland_stranger_2024} \\
$OT$ & Number of instances where a robot overtakes a group of pedestrians. &group& \cite{lu_group-aware_2025} \\
$PA$ & Average angular deviation between the pedestrians and their direct vector to their goal. &obtrusion&\cite{katyal_learning_2022} \\
$PC$ & Personal Space Cost. & comfort & \cite{kivrak_multilevel_2020} \\
$PIF$ & Passes in Front of a moving human. &comfort& \cite{che_efficient_2020} \\
$PL$ & Path Length. &path& \cite{francis2025principles} \\
$PT$ & Planning Time. & compute &\cite{shamsah_socially_2025} \\
$S$ & Success. &success& \cite{zhu_confidence-aware_2025} \\
$SC$ & Ratio of a trajectory in Space Compliance. &comfort&\cite{francis2025principles} \\
$SPL$ & Success Weighted using normalised inverse path Length.&success& \cite{francis2025principles} \\
$STL$ & Success Weighted by Time.&success& \cite{nguyen_spatiotemporal_2024} \\
$TO$ & Timeouts. &success&\cite{pohland_stranger_2024} \\
$TSC$ & Time in adhering to Space Compliance. &comfort& \cite{golchoubian_uncertainty-aware_2024} \\
$TTC$ & Estimated Minimum Time to Collision if agents' speeds remain constant. & comfort&\cite{biswas2022socnavbench} \\
\hline
\end{tabularx}
}\caption{Short names and descriptions for the metrics identified following the methodology described in Sec.~\ref{method}. We provide a reference where the metric is used and, when possible, adopt the notation in~\cite{francis2025principles}.}
\label{tab:metrics}
\end{table}
\setlength{\tabcolsep}{6pt}

\subsection*{RQ2) How frequent are surveys involving human raters?}
Out of the $85$ papers considered, $16~(18.9\%)$ papers performed in-situ human-based surveys, but only $6$ of these $(7.1\%)$ included alternative 3rd party algorithms for benchmarking purposes \cite{lu_group-aware_2025,
wang_navistar_2023, okunevich2025online, raj_rethinking_2024, hirose_sacson_2024, song_vlm-social-nav_2025}.
The remaining 10 exclusively surveyed the performance of their own proposal.
The most common questions asked in these surveys can be found in Table~\ref{tab:surveyed_metrics}.

\begin{table}[h!]
\centering
\begin{tabularx}{\columnwidth}{|X|l|}
\hline
\textbf{Variable} & \textbf{Paper} \\
\hline
Abruptness of the robot movement & \cite{navarro_sorts_2024} \\
Anxiety caused by the robot & \cite{lewandowski_socially_2020} \\
Comfort or compliance  & \cite{francis2025principles} \\
Adequacy of the perceived robot distance. & \cite{lewandowski_socially_2020} \\
Awareness and movement adequacy w.r.t. groups. & \cite{lu_group-aware_2025} \\
Overall understanding of the robot’s goals. & \cite{taylor_observer-aware_2022} \\
Likeability and friendliness & \cite{kitagawa_human-inspired_2021} \\
Naturalness and smoothness. & \cite{kruse2013human} \\
Overall navigation skills. & \cite{navarro_sorts_2024} \\
A robot's movement is easy to predict. & \cite{navarro_sorts_2024} \\
Perception of safety and risk & \cite{francis2025principles} \\
Adequacy of the robot's speed & \cite{lewandowski_socially_2020} \\
Politeness and care about pedestrians. & \cite{francis2025principles} \\
The user's trust in the robot. & \cite{che_efficient_2020} \\
Unobtrusiveness.  & \cite{hirose_sacson_2024} \\
\hline
\end{tabularx}
\caption{Variables identified as surveyed in our review, accompanied by a reference where they have been used. }
\label{tab:surveyed_metrics}
\end{table}

\subsection*{RQ3) How are benchmarking results interpreted?}
Out of 26 papers claiming better performance than the baselines used, 22 did so with either 2 or fewer metrics, or 2 or fewer algorithms, 3 of which used a single metric.
Of the same 26, 5 used a single algorithm but multiple metrics; this can be considered a reasonable to claim better performance than a specific algorithm, but not necessarily state-of-the-art performance.
Finally, 3 papers made claims that did not follow from the evidence provided. 
\par
A total of 14 papers claimed partial superiority, depending on the metrics and scenarios, and 3 papers claimed achieving results comparable but not necessarily surpassing SOTA performance.
\par
Finally, a set of 10 paper presented results with no superiority claims over the baselines, although they offered qualitative assessments of the results achieved, with respect to different metrics and/or scenarios.
\par
Overall, 18 papers acknowledged the difficulty of achieving top performance across all metrics and argued for the need to achieve a balance: 11 of them referred to balancing efficiency and comfort~\cite{martini_adaptive_2024, han_dr-mpc_2025, canh_enhancing_2024, cancelli_exploiting_2023, ansari_exploring_2023, kim_group_2022, katyal_learning_2022, linard_real-time_2023, vatan_social_2023, pohland_stranger_2024, song_vlm-social-nav_2025}, and 6 of them referred to the balance between efficiency and safety~\cite{ansari_exploring_2023, cancelli_exploiting_2023, mustafa_context_2024, zhu_confidence-aware_2025, aegidius_asfm_2024, sacco_novel_2024}.

\section{CONCLUSIONS}
Despite the efforts of the Social Robot Navigation community, benchmarking protocols are still very heterogeneous, as can be graphically appreciated in Fig.~\ref{fig:sankey}.
Although this is unsurprising when considering baselines (as new algorithms appear and older ones become obsolete), the inconsistency in the number and types of metrics used can be considered worrying.
Using a high number of metrics can be difficult to interpret, redundant if there are highly-correlated metrics, and potentially lead to biased conclusions.
However, a single metric linked to a single aspect (\textit{e.g.}, $S$ for success rate, or $SC$ for space compliance), cannot capture the complexity of social robot navigation tasks.
The number of metrics ranged from 1 (in 5 different instances) to 9 in~\cite{lu_group-aware_2025}).
Interestingly, 38 of the 85 papers reviewed, did not use any social navigation metric, and CrowdNav~\cite{liu2022intention} was the only baseline created since 2020 that the papers reviewed used as a baseline.
\par
Regarding \textbf{RQ1}:
The 4 most common \textbf{social} metrics were $DH_{min}$, $SC$, $PC$, and $C_{risk}$. Authors frequently referred to the metrics being popular, or providing good coverage of properties when making their choices, although $52.2\%$ did not provide a rationale. Most papers choose metrics such as $PL$ or $T$ over weighted metrics such as $SPL$ or $STL$, which went against our expectations.
The 4 most popular \textbf{non-social} metrics were $S$, $T$, $C$ and $PL$, with similar justifications provided as for social metrics.
The most common \textbf{baseline algorithms} used when benchmarking were ORCA~\cite{van2011reciprocal}, DWA~\cite{fox2002dynamic}, SFM~\cite{helbing1995social}, CADRL~\cite{chen2017decentralized}, SARL~\cite{chen2019crowd}, CrowdNav~\cite{liu2022intention} and LSTM-RL~\cite{everett2018motion}.
Authors frequently referred to their popularity, to their state-of-the-art status, and to baselines being the method they are improving when specifying why those were chosen---although $35.8\%$ of papers did not specify any. 
\par
Regarding \textbf{RQ2}, $18.9\%$ of the papers reviewed performed human-based surveying, but only $7.1\%$ included 3rd party algorithms in their benchmarks.
\par
Regarding \textbf{RQ3}, in most cases, papers an only claimed to have the best performance over all metrics and algorithms if the number of metrics and algorithms used for comparison were few.
A considerable number of papers, $32.7\%$, referenced directly or indirectly to the difficulty of achieving best performance in all metrics, so most papers made their claim looking only at a subset of the metrics used, or referred to social variables being more important than efficiency-related metrics such as the navigation time $T$.
\par
As a recommendation, to make benchmarking clearer, it can be beneficial to refrain from simply stating the framework used (\textit{e.g.}, ``ROS navigation stack'') when reporting results; explicitly mentioning the specific algorithm and configuration parameters can be much clearer for the readers.
\nocite{*}
\printbibliography

@article{francis2025principles,
	author = {Anthony Francis and Claudia Perez-D'Arpino and Chengshu Li and Fei Xia and Alexandre Alahi and Rachid Alami and Aniket Bera and Abhijat Biswas and Joydeep Biswas and Rohan Chandra and Hao-Tien Lewis Chiang and Michael Everett and Sehoon Ha and Justin Hart and Jonathan P. How and Haresh Karnan and Tsang-Wei Edward Lee and Luis J. Manso and Reuth Mirksy and Soeren Pirk and Phani Teja Singamaneni and Peter Stone and Ada V. Taylor and Peter Trautman and Nathan Tsoi and Marynel Vazquez and Xuesu Xiao and Peng Xu and Naoki Yokoyama and Alexander Toshev and Roberto Martin-Martin},
	title = {{Principles and Guidelines for Evaluating Social Robot Navigation Algorithms}},
	journal = {{ACM Transactions on Human-Robot Interaction}},
	year = 2025,
	volume = {{14}},
	number = {{2}},
	pages = {{65}},
	doi = {10.1145/3700599}
}

@inproceedings{lin2014microsoft,
  title={{Microsoft COCO: Common objects in context}},
  author={Lin, Tsung-Yi and Maire, Michael and Belongie, Serge and Hays, James and Perona, Pietro and Ramanan, Deva and Doll{\'a}r, Piotr and Zitnick, C Lawrence},
  booktitle={European conference on computer vision},
  pages={740--755},
  year={2014},
  organization={Springer}
}

@article{song_vlm-social-nav_2025,
	title = {{VLM}-Social-Nav: Socially Aware Robot Navigation Through Scoring Using Vision-Language Models},
	volume = {10},
	rights = {https://ieeexplore.ieee.org/Xplorehelp/downloads/license-information/{IEEE}.html},
	issn = {2377-3766, 2377-3774},
	url = {https://ieeexplore.ieee.org/document/10777573/},
	doi = {10.1109/lra.2024.3511409},
	shorttitle = {{VLM}-Social-Nav},
	pages = {508--515},
	number = {1},
	journaltitle = {{IEEE} Robotics and Automation Letters},
	shortjournal = {{IEEE} Robot. Autom. Lett.},
	author = {Song, Daeun and Liang, Jing and Payandeh, Amirreza and Raj, Amir Hossain and Xiao, Xuesu and Manocha, Dinesh},
	urldate = {2025-07-14},
	date = {2025-01},
	langid = {english},
	note = {Number: 1 Publisher: Institute of Electrical and Electronics Engineers ({IEEE})},
}

@article{golchoubian_uncertainty-aware_2024,
	title = {Uncertainty-Aware {DRL} for Autonomous Vehicle Crowd Navigation in Shared Space},
	volume = {9},
	rights = {https://ieeexplore.ieee.org/Xplorehelp/downloads/license-information/{IEEE}.html},
	issn = {2379-8904, 2379-8858},
	url = {https://ieeexplore.ieee.org/document/10538404/},
	doi = {10.1109/tiv.2024.3405330},
	pages = {7931--7944},
	number = {12},
	journaltitle = {{IEEE} Transactions on Intelligent Vehicles},
	shortjournal = {{IEEE} Trans. Intell. Veh.},
	author = {Golchoubian, Mahsa and Ghafurian, Moojan and Dautenhahn, Kerstin and Azad, Nasser Lashgarian},
	urldate = {2025-07-14},
	date = {2024-12},
	langid = {english},
	note = {Number: 12
Publisher: Institute of Electrical and Electronics Engineers ({IEEE})},
}

@inproceedings{silva_towards_2022,
	location = {Mexico City, Mexico},
	title = {Towards Online Socially Acceptable Robot Navigation},
	rights = {https://doi.org/10.15223/policy-029},
	url = {https://ieeexplore.ieee.org/document/9926686/},
	doi = {10.1109/case49997.2022.9926686},
	eventtitle = {2022 {IEEE} 18th International Conference on Automation Science and Engineering ({CASE})},
	pages = {707--714},
	booktitle = {2022 {IEEE} 18th International Conference on Automation Science and Engineering ({CASE})},
	publisher = {{IEEE}},
	author = {Silva, Steven and Paillacho, Dennys and Verdezoto, Nervo and Hernandez, Juan David},
	urldate = {2025-07-14},
	date = {2022-08-20},
	langid = {english},
}

@inproceedings{pohland_stranger_2024,
	location = {Yokohama, Japan},
	title = {Stranger Danger! Identifying and Avoiding Unpredictable Pedestrians in {RL}-based Social Robot Navigation},
	rights = {https://doi.org/10.15223/policy-029},
	url = {https://ieeexplore.ieee.org/document/10610413/},
	doi = {10.1109/icra57147.2024.10610413},
	eventtitle = {2024 {IEEE} International Conference on Robotics and Automation ({ICRA})},
	pages = {15217--15224},
	booktitle = {2024 {IEEE} International Conference on Robotics and Automation ({ICRA})},
	publisher = {{IEEE}},
	author = {Pohland, Sara and Tan, Alvin and Dutta, Prabal and Tomlin, Claire},
	urldate = {2025-07-14},
	date = {2024-05-13},
	langid = {english},
}

@inproceedings{nguyen_spatiotemporal_2024,
	location = {Ha Long, Vietnam},
	title = {Spatiotemporal Motion Profiles for Cost-Based Optimal Approaching Pose Estimation},
	rights = {https://doi.org/10.15223/policy-029},
	url = {https://ieeexplore.ieee.org/document/10417396/},
	doi = {10.1109/sii58957.2024.10417396},
	eventtitle = {2024 {IEEE}/{SICE} International Symposium on System Integration ({SII})},
	pages = {92--98},
	booktitle = {2024 {IEEE}/{SICE} International Symposium on System Integration ({SII})},
	publisher = {{IEEE}},
	author = {Nguyen, Trung-Tin and Ngo, Trung Dung},
	urldate = {2025-07-14},
	date = {2024-01-08},
	langid = {english},
}

@article{navarro_sorts_2024,
	title = {{SoRTS}: Learned Tree Search for Long Horizon Social Robot Navigation},
	volume = {9},
	rights = {https://ieeexplore.ieee.org/Xplorehelp/downloads/license-information/{IEEE}.html},
	issn = {2377-3766, 2377-3774},
	url = {https://ieeexplore.ieee.org/document/10449372/},
	doi = {10.1109/lra.2024.3370051},
	shorttitle = {{SoRTS}},
	pages = {3759--3766},
	number = {4},
	journaltitle = {{IEEE} Robotics and Automation Letters},
	shortjournal = {{IEEE} Robot. Autom. Lett.},
	author = {Navarro, Ingrid and Patrikar, Jay and Dantas, Joao P. A. and Baijal, Rohan and Higgins, Ian and Scherer, Sebastian and Oh, Jean},
	urldate = {2025-07-14},
	date = {2024-04},
	langid = {english},
	note = {Number: 4
Publisher: Institute of Electrical and Electronics Engineers ({IEEE})},
}

@inproceedings{lewandowski_socially_2020,
	location = {Naples, Italy},
	title = {Socially Compliant Human-Robot Interaction for Autonomous Scanning Tasks in Supermarket Environments},
	rights = {https://ieeexplore.ieee.org/Xplorehelp/downloads/license-information/{IEEE}.html},
	url = {https://ieeexplore.ieee.org/document/9223568/},
	doi = {10.1109/ro-man47096.2020.9223568},
	eventtitle = {2020 29th {IEEE} International Conference on Robot and Human Interactive Communication ({RO}-{MAN})},
	pages = {363--370},
	booktitle = {2020 29th {IEEE} International Conference on Robot and Human Interactive Communication ({RO}-{MAN})},
	publisher = {{IEEE}},
	author = {Lewandowski, Benjamin and Wengefeld, Tim and Muller, Sabine and Jenny, Mathias and Glende, Sebastian and Schroter, Christof and Bley, Andreas and Gross, Horst-Michael},
	urldate = {2025-07-14},
	date = {2020-08},
	langid = {english},
}

@article{shamsah_socially_2025,
	title = {Socially Acceptable Bipedal Robot Navigation via Social Zonotope Network Model Predictive Control},
	volume = {22},
	rights = {https://ieeexplore.ieee.org/Xplorehelp/downloads/license-information/{IEEE}.html},
	issn = {1545-5955, 1558-3783},
	url = {https://ieeexplore.ieee.org/document/10810741/},
	doi = {10.1109/tase.2024.3519012},
	pages = {10130--10148},
	journaltitle = {{IEEE} Transactions on Automation Science and Engineering},
	shortjournal = {{IEEE} Trans. Automat. Sci. Eng.},
	author = {Shamsah, Abdulaziz and Agarwal, Krishanu and Katta, Nigam and Raju, Abirath and Kousik, Shreyas and Zhao, Ye},
	urldate = {2025-07-14},
	date = {2025},
	langid = {english},
	note = {Publisher: Institute of Electrical and Electronics Engineers ({IEEE})},
}

@inproceedings{vatan_social_2023,
	location = {Coimbra, Portugal},
	title = {Social {APF}-{RL}: Safe Mapless Navigation in Unknown \& Human-Populated Environments},
	rights = {https://doi.org/10.15223/policy-029},
	url = {https://ieeexplore.ieee.org/document/10256274/},
	doi = {10.1109/ecmr59166.2023.10256274},
	shorttitle = {Social {APF}-{RL}},
	eventtitle = {2023 European Conference on Mobile Robots ({ECMR})},
	pages = {1--6},
	booktitle = {2023 European Conference on Mobile Robots ({ECMR})},
	publisher = {{IEEE}},
	author = {Vatan, S. Batuhan and Bektaş, Kemal and Bozma, H. Işıl},
	urldate = {2025-07-14},
	date = {2023-09-04},
	langid = {english},
}

@article{hirose_sacson_2024,
	title = {{SACSoN}: Scalable Autonomous Control for Social Navigation},
	volume = {9},
	rights = {https://ieeexplore.ieee.org/Xplorehelp/downloads/license-information/{IEEE}.html},
	issn = {2377-3766, 2377-3774},
	url = {https://ieeexplore.ieee.org/document/10305270/},
	doi = {10.1109/lra.2023.3329626},
	shorttitle = {{SACSoN}},
	pages = {49--56},
	number = {1},
	journaltitle = {{IEEE} Robotics and Automation Letters},
	shortjournal = {{IEEE} Robot. Autom. Lett.},
	author = {Hirose, Noriaki and Shah, Dhruv and Sridhar, Ajay and Levine, Sergey},
	urldate = {2025-07-14},
	date = {2024-01},
	langid = {english},
	note = {Number: 1
Publisher: Institute of Electrical and Electronics Engineers ({IEEE})},
}

@inproceedings{raj_rethinking_2024,
	location = {Yokohama, Japan},
	title = {Rethinking Social Robot Navigation: Leveraging the Best of Two Worlds},
	rights = {https://doi.org/10.15223/policy-029},
	url = {https://ieeexplore.ieee.org/document/10611710/},
	doi = {10.1109/icra57147.2024.10611710},
	shorttitle = {Rethinking Social Robot Navigation},
	eventtitle = {2024 {IEEE} International Conference on Robotics and Automation ({ICRA})},
	pages = {16330--16337},
	booktitle = {2024 {IEEE} International Conference on Robotics and Automation ({ICRA})},
	publisher = {{IEEE}},
	author = {Raj, Amir Hossain and Hu, Zichao and Karnan, Haresh and Chandra, Rohan and Payandeh, Amirreza and Mao, Luisa and Stone, Peter and Biswas, Joydeep and Xiao, Xuesu},
	urldate = {2025-07-14},
	date = {2024-05-13},
	langid = {english},
}

@inproceedings{linard_real-time_2023,
	location = {Detroit, {MI}, {USA}},
	title = {Real-Time {RRT}$^{\textrm{*}}$ with Signal Temporal Logic Preferences},
	rights = {https://doi.org/10.15223/policy-029},
	url = {https://ieeexplore.ieee.org/document/10341993/},
	doi = {10.1109/iros55552.2023.10341993},
	eventtitle = {2023 {IEEE}/{RSJ} International Conference on Intelligent Robots and Systems ({IROS})},
	pages = {8621--8627},
	booktitle = {2023 {IEEE}/{RSJ} International Conference on Intelligent Robots and Systems ({IROS})},
	publisher = {{IEEE}},
	author = {Linard, Alexis and Torre, Ilaria and Bartoli, Ermanno and Sleat, Alex and Leite, Iolanda and Tumova, Jana},
	urldate = {2025-07-14},
	date = {2023-10-01},
	langid = {english},
}

@inproceedings{shamsah_real-time_2024,
	location = {Abu Dhabi, United Arab Emirates},
	title = {Real-time Model Predictive Control with Zonotope-Based Neural Networks for Bipedal Social Navigation},
	rights = {https://doi.org/10.15223/policy-029},
	url = {https://ieeexplore.ieee.org/document/10801435/},
	doi = {10.1109/iros58592.2024.10801435},
	eventtitle = {2024 {IEEE}/{RSJ} International Conference on Intelligent Robots and Systems ({IROS})},
	pages = {13741--13748},
	booktitle = {2024 {IEEE}/{RSJ} International Conference on Intelligent Robots and Systems ({IROS})},
	publisher = {{IEEE}},
	author = {Shamsah, Abdulaziz and Agarwal, Krishanu and Kousik, Shreyas and Zhao, Ye},
	urldate = {2025-07-14},
	date = {2024-10-14},
	langid = {english},
}

@inproceedings{taylor_observer-aware_2022,
	location = {Napoli, Italy},
	title = {Observer-Aware Legibility for Social Navigation},
	rights = {https://doi.org/10.15223/policy-029},
	url = {https://ieeexplore.ieee.org/document/9900676/},
	doi = {10.1109/ro-man53752.2022.9900676},
	eventtitle = {2022 31st {IEEE} International Conference on Robot and Human Interactive Communication ({RO}-{MAN})},
	pages = {1115--1122},
	booktitle = {2022 31st {IEEE} International Conference on Robot and Human Interactive Communication ({RO}-{MAN})},
	publisher = {{IEEE}},
	author = {Taylor, Ada V. and Mamantov, Ellie and Admoni, Henny},
	urldate = {2025-07-14},
	date = {2022-08-29},
	langid = {english},
}

@inproceedings{wang_navistar_2023,
	location = {Detroit, {MI}, {USA}},
	title = {{NaviSTAR}: Socially Aware Robot Navigation with Hybrid Spatio-Temporal Graph Transformer and Preference Learning},
	rights = {https://doi.org/10.15223/policy-029},
	url = {https://ieeexplore.ieee.org/document/10341395/},
	doi = {10.1109/iros55552.2023.10341395},
	shorttitle = {{NaviSTAR}},
	eventtitle = {2023 {IEEE}/{RSJ} International Conference on Intelligent Robots and Systems ({IROS})},
	pages = {11348--11355},
	booktitle = {2023 {IEEE}/{RSJ} International Conference on Intelligent Robots and Systems ({IROS})},
	publisher = {{IEEE}},
	author = {Wang, Weizheng and Wang, Ruiqi and Mao, Le and Min, Byung-Cheol},
	urldate = {2025-07-14},
	date = {2023-10-01},
	langid = {english},
}

@inproceedings{katyal_learning_2022,
	location = {Kyoto, Japan},
	title = {Learning a Group-Aware Policy for Robot Navigation},
	rights = {https://doi.org/10.15223/policy-029},
	url = {https://ieeexplore.ieee.org/document/9981183/},
	doi = {10.1109/iros47612.2022.9981183},
	eventtitle = {2022 {IEEE}/{RSJ} International Conference on Intelligent Robots and Systems ({IROS})},
	pages = {11328--11335},
	booktitle = {2022 {IEEE}/{RSJ} International Conference on Intelligent Robots and Systems ({IROS})},
	publisher = {{IEEE}},
	author = {Katyal, Kapil and Gao, Yuxiang and Markowitz, Jared and Pohland, Sara and Rivera, Corban and Wang, I-Jeng and Huang, Chien-Ming},
	urldate = {2025-07-14},
	date = {2022-10-23},
	langid = {english},
}

@inproceedings{kitagawa_human-inspired_2021,
	location = {Boulder {CO} {USA}},
	title = {Human-inspired Motion Planning for Omni-directional Social Robots},
	rights = {https://www.acm.org/publications/policies/copyright\_policy\#Background},
	url = {https://dl.acm.org/doi/10.1145/3434073.3444679},
	doi = {10.1145/3434073.3444679},
	eventtitle = {{HRI} '21: {ACM}/{IEEE} International Conference on Human-Robot Interaction},
	pages = {34--42},
	booktitle = {Proceedings of the 2021 {ACM}/{IEEE} International Conference on Human-Robot Interaction},
	publisher = {{ACM}},
	author = {Kitagawa, Ryo and Liu, Yuyi and Kanda, Takayuki},
	urldate = {2025-07-14},
	date = {2021-03-08},
	langid = {english},
}

@inproceedings{kim_group_2022,
	location = {Jeju, Korea, Republic of},
	title = {Group Estimation for Social Robot Navigation in Crowded Environments},
	rights = {https://doi.org/10.15223/policy-029},
	url = {https://ieeexplore.ieee.org/document/10003761/},
	doi = {10.23919/iccas55662.2022.10003761},
	eventtitle = {2022 22nd International Conference on Control, Automation and Systems ({ICCAS})},
	pages = {1421--1425},
	booktitle = {2022 22nd International Conference on Control, Automation and Systems ({ICCAS})},
	publisher = {{IEEE}},
	author = {Kim, Mincheul and Kwon, Youngsun and Yoon, Sung-Eui},
	urldate = {2025-07-14},
	date = {2022-11-27},
	langid = {english},
}

@article{lu_group-aware_2025,
	title = {Group-Aware Robot Navigation in Crowds Using Spatio-Temporal Graph Attention Network With Deep Reinforcement Learning},
	volume = {10},
	rights = {https://ieeexplore.ieee.org/Xplorehelp/downloads/license-information/{IEEE}.html},
	issn = {2377-3766, 2377-3774},
	url = {https://ieeexplore.ieee.org/document/10918817/},
	doi = {10.1109/lra.2025.3549663},
	pages = {4140--4147},
	number = {4},
	journaltitle = {{IEEE} Robotics and Automation Letters},
	shortjournal = {{IEEE} Robot. Autom. Lett.},
	author = {Lu, Xiaojun and Faragasso, Angela and Wang, Yongdong and Yamashita, Atsushi and Asama, Hajime},
	urldate = {2025-07-14},
	date = {2025-04},
	langid = {english},
	note = {Number: 4
Publisher: Institute of Electrical and Electronics Engineers ({IEEE})},
}

@inproceedings{ansari_exploring_2023,
	location = {Detroit, {MI}, {USA}},
	title = {Exploring Social Motion Latent Space and Human Awareness for Effective Robot Navigation in Crowded Environments},
	rights = {https://doi.org/10.15223/policy-029},
	url = {https://ieeexplore.ieee.org/document/10341721/},
	doi = {10.1109/iros55552.2023.10341721},
	eventtitle = {2023 {IEEE}/{RSJ} International Conference on Intelligent Robots and Systems ({IROS})},
	pages = {1--8},
	booktitle = {2023 {IEEE}/{RSJ} International Conference on Intelligent Robots and Systems ({IROS})},
	publisher = {{IEEE}},
	author = {Ansari, Junaid Ahmed and Tourani, Satyajit and Kumar, Gourav and Bhowmick, Brojeshwar},
	urldate = {2025-07-14},
	date = {2023-10-01},
	langid = {english},
}

@inproceedings{cancelli_exploiting_2023,
	location = {Paris, France},
	title = {Exploiting Proximity-Aware Tasks for Embodied Social Navigation},
	rights = {https://doi.org/10.15223/policy-029},
	url = {https://ieeexplore.ieee.org/document/10377162/},
	doi = {10.1109/iccv51070.2023.01006},
	eventtitle = {2023 {IEEE}/{CVF} International Conference on Computer Vision ({ICCV})},
	pages = {10923--10933},
	booktitle = {2023 {IEEE}/{CVF} International Conference on Computer Vision ({ICCV})},
	publisher = {{IEEE}},
	author = {Cancelli, Enrico and Campari, Tommaso and Serafini, Luciano and Chang, Angel X. and Ballan, Lamberto},
	urldate = {2025-07-14},
	date = {2023-10-01},
	langid = {english},
}

@inproceedings{canh_enhancing_2024,
	location = {Jeju, Korea, Republic of},
	title = {Enhancing Social Robot Navigation with Integrated Motion Prediction and Trajectory Planning in Dynamic Human Environments},
	rights = {https://doi.org/10.15223/policy-029},
	url = {https://ieeexplore.ieee.org/document/10773156/},
	doi = {10.23919/iccas63016.2024.10773156},
	eventtitle = {2024 24th International Conference on Control, Automation and Systems ({ICCAS})},
	pages = {731--736},
	booktitle = {2024 24th International Conference on Control, Automation and Systems ({ICCAS})},
	publisher = {{IEEE}},
	author = {Canh, Thanh Nguyen and {HoangVan}, Xiem and Chong, Nak Young},
	urldate = {2025-07-14},
	date = {2024-10-29},
	langid = {english},
}

@article{che_efficient_2020,
	title = {Efficient and Trustworthy Social Navigation via Explicit and Implicit Robot-Human Communication},
	volume = {36},
	rights = {https://ieeexplore.ieee.org/Xplorehelp/downloads/license-information/{IEEE}.html},
	issn = {1552-3098, 1941-0468},
	url = {https://ieeexplore.ieee.org/document/8967120/},
	doi = {10.1109/tro.2020.2964824},
	pages = {692--707},
	number = {3},
	journaltitle = {{IEEE} Transactions on Robotics},
	shortjournal = {{IEEE} Trans. Robot.},
	author = {Che, Yuhang and Okamura, Allison M. and Sadigh, Dorsa},
	urldate = {2025-07-14},
	date = {2020-06},
	langid = {english},
	note = {Number: 3
Publisher: Institute of Electrical and Electronics Engineers ({IEEE})},
}

@article{han_dr-mpc_2025,
	title = {{DR}-{MPC}: Deep Residual Model Predictive Control for Real-World Social Navigation},
	volume = {10},
	rights = {https://ieeexplore.ieee.org/Xplorehelp/downloads/license-information/{IEEE}.html},
	issn = {2377-3766, 2377-3774},
	url = {https://ieeexplore.ieee.org/document/10904316/},
	doi = {10.1109/lra.2025.3546106},
	shorttitle = {{DR}-{MPC}},
	pages = {4029--4036},
	number = {4},
	journaltitle = {{IEEE} Robotics and Automation Letters},
	shortjournal = {{IEEE} Robot. Autom. Lett.},
	author = {Han, James R. and Thomas, Hugues and Zhang, Jian and Rhinehart, Nicholas and Barfoot, Timothy D.},
	urldate = {2025-07-14},
	date = {2025-04},
	langid = {english},
	note = {Number: 4
Publisher: Institute of Electrical and Electronics Engineers ({IEEE})},
}

@inproceedings{mustafa_context_2024,
	location = {London, United Kingdom},
	title = {Context Aware Mamba-based Reinforcement Learning for Social Robot Navigation},
	rights = {https://doi.org/10.15223/policy-029},
	url = {https://ieeexplore.ieee.org/document/10843924/},
	doi = {10.1109/iccma63715.2024.10843924},
	eventtitle = {2024 12th International Conference on Control, Mechatronics and Automation ({ICCMA})},
	pages = {154--159},
	booktitle = {2024 12th International Conference on Control, Mechatronics and Automation ({ICCMA})},
	publisher = {{IEEE}},
	author = {Mustafa, Syed Muhammad and Ahmed Usmani, Zain and Rizvi, Omema and Memon, Abdul Basit and Mobeen Movania, Muhammad},
	urldate = {2025-07-14},
	date = {2024-11-11},
	langid = {english},
}

@inproceedings{van2011reciprocal,
  title={Reciprocal n-body collision avoidance},
  author={Van Den Berg, Jur and Guy, Stephen J and Lin, Ming and Manocha, Dinesh},
  booktitle={Robotics Research: The 14th International Symposium ISRR},
  pages={3--19},
  year={2011},
  organization={Springer}
}
\end{document}